\documentclass{article}

\usepackage{graphicx}
\usepackage{subfigure}
\usepackage{booktabs} 
\usepackage{bbm}
\usepackage{ragged2e}
\usepackage{hyperref}
\usepackage{amsfonts}
\usepackage{siunitx}
\usepackage{dcolumn}
\usepackage{microtype}

\usepackage{siunitx}
\usepackage[accepted]{icml2024}

\usepackage{amsmath}
\usepackage{amssymb}
\usepackage{mathtools}
\usepackage{microtype}
\usepackage{amsthm}
\usepackage[linesnumbered,ruled,vlined, algo2e]{algorithm2e}
\usepackage{
    amsmath,
    algorithm,
    float,
    tabularx,
    amsfonts,
}
\makeatletter
\newcommand{\multiline}[1]{%
    \begin{tabularx}{\dimexpr\linewidth-\ALG@thistlm}[t]{@{}X@{}}
        #1
    \end{tabularx}
}
\newcommand{\Input}[1]{\algrenewcommand{\alglinenumber}[1]{Input: \ \setcounter{ALG@line}{\numexpr##1-1}} #1}

\newcommand{\Output}[1]{\algrenewcommand{\alglinenumber}[1]{Output:\setcounter{ALG@line}{\numexpr##1-1}} #1}

\usepackage[capitalize,noabbrev]{cleveref}
\usepackage{tikz,lipsum,lmodern}
\usepackage[most]{tcolorbox}

\theoremstyle{plain}

\theoremstyle{definition}

\theoremstyle{remark}

\usepackage[textsize=tiny]{todonotes}

\icmltitlerunning{Learning Beyond Pattern Matching?}

\begin{document}

\twocolumn[
\icmltitle{Learning Beyond Pattern Matching? \\
Assaying Mathematical Understanding in LLMs}



\icmlsetsymbol{equal}{*}

\begin{icmlauthorlist}
\icmlauthor{Siyuan Guo}{mpi,cam}
\icmlauthor{Aniket Didolkar}{mila}
\icmlauthor{Nan Rosemary Ke}{google}
\\
\icmlauthor{Anirudh Goyal}{mila,equal}
\icmlauthor{Ferenc Huszár}{cam,equal}
\icmlauthor{Bernhard Schölkopf}{mpi,equal}
\end{icmlauthorlist}

\icmlaffiliation{cam}{Department of Computer Science, University of Cambridge, United Kingdom}
\icmlaffiliation{google}{Google DeepMind}
\icmlaffiliation{mpi}{Max Planck Institute for Intelligent Systems, Tübingen, Germany}
\icmlaffiliation{mila}{Mila, University of Montreal}

\icmlcorrespondingauthor{Siyuan Guo}{siyuan.guo@tuebingen.mpg.de}

\icmlkeywords{Machine Learning, ICML}

\vskip 0.3in
]



\printAffiliationsAndNotice{*Joint senior authors, listed alphabetically}  

\begin{abstract}
We are beginning to see progress in language model assisted scientific discovery. Motivated by the use of LLMs as a general scientific assistant, this paper assesses the domain knowledge of LLMs through its understanding of different mathematical skills required to solve problems. In particular, we look at not just what the pre-trained model already knows, but how it learned to learn from information during in-context learning or instruction-tuning through exploiting the complex knowledge structure within mathematics. Motivated by the Neural Tangent Kernel (NTK), we propose \textit{NTKEval} to assess changes in LLM's probability distribution via training on different kinds of math data. Our systematic analysis finds evidence of domain understanding during in-context learning. By contrast, certain instruction-tuning leads to similar performance changes irrespective of training on different data, suggesting a lack of domain understanding across different skills.  \looseness=-1
\end{abstract}

\begin{figure*}[h]
    
\begin{minipage}{.49\textwidth}
\begin{tcolorbox}[colback=gray!5!white,colframe=gray!75!black,title=Example questions: elementary to complex skills]
\vspace{-0.5cm}
\begin{align*}
    \text{ADD:} \quad &\text{What is }41 + 55? \ \text{A: }96\\
    \text{SUB:} \quad &\text{What is }450 - 597? \ \text{A: } -147\\
    \text{MUL:} \quad &\text{What is }9 * 50? \ \text{A: }450\\
    \text{DIV:} \quad &\text{What is }4410 / 63? \ \text{A: }70\\
    \text{OPS:} \quad &\text{What is }(33 / 33) * 64? \ \text{A: }64 \\
    \text{CPLEX:} \quad &\text{What is }28*(49 + (58 - 52))+134\\
    &+((126 / 21) + 106)?\ \text{A: }1786
\end{align*}
\end{tcolorbox}
\end{minipage}
\hspace{0.3cm}
\begin{minipage}{.49\textwidth}
\begin{tcolorbox}[colback=gray!5!white,colframe=gray!75!black,title=Example questions: surface changes in formats]
\vspace{-0.5cm}
\begin{align*}
    \text{question:} \quad & \text{What is }190 - 8? \ \text{A: }182 \\
    \text{instruction:} \quad & \text{Subtract } 8 \text{ from } 190 \ \text{A: }182 \\
    \text{symbolic:} \quad & \text{A}=190, \text{B}=8. \ \text{A}- \text{B} =? \\
    &\text{A: } 182 \\
    \text{word problem:} \quad & \text{Amy made 190 dollars profit and} \\
    & \text{spends 8 dollars as her cost.} \\
    & \text{What is her net profit?} \ \text{A: }182
\end{align*}
\end{tcolorbox}
\end{minipage}
\caption{Example questions from synthetic dataset. Left shows math skill progression from elementary to complex skills and right shows a list of presentation formats (i.e. surface structures) while the question tests the same deep math skill. \looseness=-1}
\label{fig:sample_synthetic_questions}
\end{figure*}

\section{Introduction}

Large Language Models (LLMs) have demonstrated remarkable success in diverse natural language inference tasks \cite{achiam2023gpt,touvron2023llama,chowdhery2023palm,anil2023palm,touvron2023llama2,team2023gemini}. 
With the promising success, there is a growing trend to use LLM as creative assistants for scientific discovery: from mathematics \cite{trinh_solving_2024} to biology \cite{madani_large_2023}.   

Motivated by the use of LLM as a scientific assistant, our paper assesses the domain knowledge of LLMs through their understanding of different mathematical skills required to solve problems. Understanding can be measured in two ways: the degree to which it solves problems correctly; and the degree to which it enables fast adaptation to new knowledge. Similarly, ``understanding" in an LLM has two facets: on the one hand, pre-trained LLMs possess knowledge that allows remarkable performance in zero-shot tasks; on the other hand, pre-trained LLMs can learn new knowledge, either by leveraging in-context learning or by instruction-tuning from base parameters as initialization. While most evaluation focuses on measuring what the model knows already, we focus on evaluating an LLM's mathematical understanding by studying how they learn: Has the model \emph{learnt to learn} effectively about mathematics? Can it make good use of relevant information present during learning? In this sense, our interest is in evaluating LLMs from a learning-to-learn, or meta-learning perspective.

\looseness=-1 


Humans successfully acquire mathematical and reasoning skills when they are able to identify underlying structure in problems rather than paying attention to spurious signals present in the question formulation --- a phenomenon often coined as deep versus surface learning in education science \cite{chindeep2000}. This suggests that an analogous way to evaluate a machine learning model's understanding of mathematics is to ask whether it is able to exploit similarities in deep structure as it generalizes from a training example to a test situation. In gradient-based learning, the extent to which information about one input is generalized to another is captured by the object called Neural Tangent Kernel (NTK) \cite{jacot2018neural}. Although the NTK was mainly introduced as a theoretical tool, it has also found uses in interpretability research \cite{Engel2023faithful}. \looseness=-1

This paper proposes a NTK-inspired method to evaluate LLM's change in probability distribution during training and investigates: do LLMs learn to answer math problems based on an understanding of the skill (deep structure) required to solve the problem or by gathering clues from surface changes in presentation formats? For example, to solve a subtraction problem, LLMs' performance increase may be due to noticing the train and test question shares the same symbolic format ($A=825,B=192$, what is $A-B$?) instead of eliciting the knowledge of subtraction for problem-solving. 
Figure \ref{fig:sample_synthetic_questions} shows more examples of different presentation formats considered. \looseness=-1

We assess the impact on LLM's accuracy in answering math problems when LLMs see different groups of examples that relate to the test question: one group shares the same deep structure described by core math skill, and the other group shares the same presentation format. This is based on the intuition that if LLMs learn beyond pattern matching, then seeing deep structures should induce larger relative improvement than seeing surface structures. Subsequently, we analyze, for a fixed presentation format, how seeing different math skills affects performance on targeted and different test examples. For instance, we measure accuracy improvement (or decline) in how the LLM sees addition examples affect solving addition problems versus subtraction problems, to determine  LLM's ability in fast adaptation. 
In-context learning \cite{brown2020language} and instruction-tuning \cite{zhou2023lima} have elicited emergent abilities of LLMs: from improvements in reasoning \cite{wei2022emergent, wei2022chain} to generalization beyond training dataset \cite{wei2021finetuned}. In this paper, we analyze from the two perspectives and our contributions and findings are summarized below:

\vspace{-0.3cm}
\begin{itemize}

    \item We propose \textit{NTKEval} in Section \ref{sec:NTKEval}, extending NTK to language models when outputs are chat completions, and demonstrate the sample efficiency of \textit{NTKEval} compared to standard metric in counting accuracy differences in Section \ref{sec:NTKEval_sample_efficiency}. 
    
    \item We introduce the KhanSkill dataset (Section \ref{sec:datasets}) consisting of human-annotated mathematical concepts that can help analyze the alignment of LLMs' mathematical understanding with human learning.
    
    \item Our systematic analysis in Section \ref{sec:deep_shallow} and \ref{sec:elementary_to_complex} finds that in-context learning differentiates deep versus surface structures (Table \ref{table:ICL_deep_shallow}) and learned to effectively use relevant math skills (Figure \ref{fig:targetd_vs_off_diagonals} Top), whereas instruction-tuning on skill-focused dataset leads to similar performance change irrespective of training on different data types (Table \ref{table:FT_deep_shallow} and Figure \ref{fig:targetd_vs_off_diagonals} Bottom) --- suggesting the adaptation is on format matching rather than domain understanding.
\end{itemize}

\section{Related Work}
\textbf{LLMs for Math problem solving} Current breakthroughs, e.g., OPRO \cite{yang2023large}, AlphaGeometry \cite{trinh_solving_2024}, FunSearch \cite{romera-paredes_mathematical_2024},  in AI for mathematics were driven by the idea of self-play where LLM acts as a creative assistant to provide useful hints. The advantages brought by language model relies more on its general understanding across domain concepts rather than its ability to precisely execute manual tasks --- which is often outsourced to reliable external tools \cite{reed2022generalist, schick2023toolformer}. This motivates our study to assess the domain understanding of LLMs through their ability to adapt when seeing different math skills.

\textbf{Neural Tangent Kernel} \cite{jacot2018neural} is central to understanding the generalization properties of ANNs --- it shows with infinite width network, the kernel is deterministic with respect to model architecture rather than parameter initialization and stays constant during training \cite{weng2022ntk}. Much of the existing literature \cite{bietti2019inductive, alemohammad2020recurrent, chen2020generalized} focuses on analyzing the theoretical properties of NTK in various architectures, with little connections to language models. A more related work \cite{malladi2023kernel} studies whether NTK describes the fine-tuning process of language models where they formulate downstream tasks as masked word prediction problems through prompting. In this paper, we instead propose a sample-efficient method that allows outputs to be free-form completions. We utilize the proposed method to study how effectively models can learn through training on relevant data.

\textbf{Skill} 
\citet{arora2023theory} studies skill emergence in language models from a statistical framework.  \citet{chen2023skillit} selects training data based on skill ordering. \citet{chen2023skills} introduces SkiC prompting to encourage skill compositions. 

\section{Background}
\subsection{Language Model}
A language model (LM) is a statistical model of natural language. Given a sentence $s$ with tokens $w_1, \ldots, w_t$, the probability of the sentence in a LM parameterized by $\theta$ can be represented as a chain of conditional probability conditioned on all the previous tokens:
\begin{align}
    p_\theta(s) = \prod_{i=1}^t p_\theta(w_i \mid w_1, \ldots, w_{i-1})
\end{align}
Pre-training a language model amounts to likelihood maximization for the probability of next token prediction $p_\theta(w_t | w_1, \ldots, w_{t-1})$ on a held-out dataset. 


\textbf{In-context learning} LM is given a $k$-shot example of context at inference time before the context of the test question, which the LM is expected to complete. \looseness=-1

\textbf{Instruction-tuning} Given a dataset of questions and ground truth answers $\{(\mathbf{x}_i, \mathbf{y}_i)\}$ (such dataset is what this paper focuses to analyze) and denoting the tokens in $i$-th question as $\mathbf{x}_i := [x_{i,1}, x_{i,2}, \ldots]$ and in the corresponding answer as $\mathbf{y}_i := [y_{i,1}, \ldots, y_{i, T_i}]$. Instruction-tuning or supervised fine-tuning a LM $p_\theta$ on the dataset means minimizing the loss function 
\begin{align*}
    L(\theta) = - \sum_i &\bigg [\sum_{t=1}^{T_i - 1} \log p_\theta(y_{i, t+1} | y_{i,t:1}, \mathbf{x}_i)\\& + \log p_\theta(y_{i, 1} | \mathbf{x}_i) \bigg] 
\end{align*}

\subsection{Neural Tangent Kernel}
Given a neural network $f$ with parameter $\theta$, for an input pair $x, x'$, NTK quantifies the changes in $f_\theta$ at point $x$ when updating an infinitesimal gradient step in the direction of training on data point $x'$. Mathematically, the kernel can be written in two equivalent forms, where $\eta$ is the learning rate:
\begin{align}
    k(x, x') & = \lim_{\eta \to 0} \frac{f(x; \theta + \eta \frac{\partial f_\theta(x')}{\partial \theta}) - f(x; \theta)}{\eta} \label{ntkeq} \\
    &= \nabla_\theta f(x; \theta)^T \nabla_\theta f(x'; \theta)
\end{align}
In this work, we inherit Eq.~\eqref{ntkeq} for kernel calculation since language models usually have parameters in the scale of (tens of) billions, and matrix multiplications with billion entries are computationally expensive.

\section{Methods}
\label{sec:NTKEval}
This paper proposes a NTK-inspired method \textit{NTKEval} to evaluate changes in probability when training on a skill-focused dataset. We expect changes in probability would be faster in capturing training effects that are otherwise computationally expensive to reflect in accuracy changes. We focus on analyzing math problems with deterministic solutions. Each data point consists of a triplet of skill, question, and answer, where skill describes a feature of the question and the answer may contain CoT reasoning \cite{wei2022chain} where an exact solution is extractable. Let $\mathbf{x}_s, \mathbf{x}_{s'}$ represent batches of data points corresponding to skill, $s$ and $s'$ respectively. Here a skill can either be a math skill (deep structure) or a type of presentation format (surface structure). Our goal is to learn the language model's NTK in terms of skills, $k(s, s') := k(\mathbf{x}_s, \mathbf{x}_{s'})$. 

Computing neural tangent kernel in language models faces three challenges: \looseness=-1

NTK is designed to tackle regression tasks, where $f_\theta(\cdot)$ is a numerical value. Language models instead generate free-form completions based on a prompt. To address the lack of target numerical outcome, we use the probability of generating the correct solution given prompt, i.e., $p_\theta(\text{correct} \mid \text{prompt})$ as the target value. 

To compute $p_\theta(\text{correct} \mid \text{prompt})$, we marginalize the CoT reasoning effect. Formally, given a prompt $x$, the model generates $i$-th completion $c_i = (z_i, y_i)$, where $z_i$ is the CoT reasoning and $y_i$ is the deterministic solution. Let the ground truth solution be $y$. Then, 
\begin{align}
&p_\theta(\text{correct} \mid \text{prompt}) = p_\theta(y \mid x) \\
&= \frac{1}{N}\sum_{(z_i, y_i) \sim p_\theta(\cdot | x)} p_\theta(z_i, y_i |x) \mathbbm{1}_{y_i = y},
\end{align}
where $N$ is the number of generations given prompt $x$. 

However, language models are known to have high variability in free-form generations when sampling from the complete probability distribution. Therefore, sampling multiple generations in each model and marginalizing out chain-of-thought reasoning, though theoretically correct, in practice will be computationally expensive as the number of generations required is huge. Here we take the importance sampling approach: use the generations in the base model and calculate the counterfactual probability of the new model generating the same completion. This allows fair comparisons of the same inputs between different models and requires fewer computation resources. Formally, let model $0$ be the base model and model $1$ be the instruction-tuned model on a data batch that shares the same skill, with probability distributions $p_0, p_1$ respectively. The objective is to calculate the difference in probability of generating correct solutions between model $0$ and $1$ given fixed prompt $x$, i.e., $p_1(y | x) - p_0(y|x)$. Importance sampling based score leads to 
\begin{align}
    p_1(y | x) - p_0(y|x) = \frac{1}{N} \sum_{c_i \sim p_0(\cdot | x)} \big[ \frac{p_1(c_i|x)}{p_0(c_i | x)} - 1\big] s(y_i, y), 
\end{align}
where $s(y_i, y) = 1 \text{ if } y = y_i \text{ else } -1$. Note we adapt the indicator function slightly to take into account the changes when model outputs wrong answers. Algorithm \ref{algorithm:NTKEval} details the exact procedure. 
\begin{algorithm}[h]
\caption{NTKEval: Algorithm to compute neural tangent kernel when training on skill-focused dataset}

\label{algorithm:NTKEval}
\SetKwInOut{Input}{Input}
\SetKwInOut{Output}{Output}    
    \Input{Base LLM: Model 0 \\
    Training batch $\mathcal{D}_s := (\mathbf{x}_s, \mathbf{y}_s)$ on skill $s$ \\
    Evaluation batch $\mathcal{D}_{s'} := (\mathbf{x}_{s'}, \mathbf{y}_{s'})$ on skill $s'$ \\
    Number of generations per prompt: $N$}
    \Output{Neural tangent kernel $k(s, s') := k(\mathcal{D}_s, \mathcal{D}_{s'})$}
    
    Instruction-tune model $0$ on $\mathcal{D}_s$ to get model 1 \;
    
    \For{each question answer pair $(x_s, y_s)$ in $\mathcal{D}_s$}
    {
    Concat fixed prompt with target question $x_s$ \;
    
    Generate $N$ completions given prompt under model $0$, where the completions are denoted as $\{c_i\}_{i=1}^N$. \;

        \For{each $c_i$}
        {
        Extract deterministic solution $y_i$ from $c_i$ and compute the indicator function $s(y_i, y_s)$ \;
        
        Compute the log probability of Model $0$ in generating $c_i$ given question $x_s$ : $\log p_0(c_i|x_s)$ \; 
    
        Compute the log probability of Model $1$ in generating $c_i$ given question $x_s$: $\log p_1(c_i | x_s)$ \;
        }
    }
    \textbf{Return} $k(s, s')$ as in 
    \begin{align*}
    \frac{1}{\lvert \mathcal{D}_s \rvert }\sum_{(x_s, y_s) \in \mathcal{D}_s} \bigg[\frac{1}{N} \sum_{c_i \sim p_0(\cdot | x_s)} \big[ \frac{p_1(c_i|x_s)}{p_0(c_i | x_s)} - 1\big] s(y_i, y_s) \bigg],
    \end{align*}
    where $s'$ is implicit under training to reach Model $1$. 

\end{algorithm}

\section{Datasets}
\label{sec:datasets}
\textbf{Synthetic dataset} To allow control in isolating changes on variables of interest, this paper focuses the analysis on synthetic datasets. Each question tests one math skill. We first create one dataset consisting of questions testing four elementary math skills (`deep structure') and presented with four types of formats (`surface structure'). The four elementary skills are addition, subtraction, multiplication, and division and the four types of presentation formats are question, instruction, symbolic, and word problems. Figure \ref{fig:sample_synthetic_questions} (right) shows example formats for a given math problem. 

To understand whether LMs are effective in learning to learn through seeing different math skills, we fix the presentation format to style `question' and create the second dataset with questions that require understanding from elementary to more complex skills. In particular, we add questions testing understanding on the order of operations, and the utilization of the mixture of all aforementioned skills (`complex'). Figure \ref{fig:sample_synthetic_questions} (left) shows example questions with abbreviated skill names. Finally, we include a baseline that outputs random integers irrespective of the question presented.

\textbf{KhanSkill dataset}
Due to a lack of skill-annotated benchmarks, we generate expert-written questions for educational purposes from \textit{khan-exercises}\footnote{https://github.com/Khan/khan-exercises} that reflect human understanding. The dataset consists of $93$ skills with $20$ questions per skill. The training set consists of $1240$ questions and the test set consists of $620$ questions evenly split across skills. Appendix \ref{sec:appendix_dataset} details sample skills and questions contained in the dataset. 

\begin{table}
\centering
\begin{footnotesize}
\begin{sc}
\begin{tabular}{lcccc}
\toprule
Model & rand & add & sub & mul \\
\midrule
Codellama-7b  & 0.01 & 0.87 & 0.71 & 0.19  \\
Llemma-7b & 0.01 & 0.98 & 0.94 & 0.44 \\
Mistral-7b & 0.01 & 0.99 & 0.97 & 0.51 \\
Mixtral-8x7b-Instruct & 0.01 & 0.98 & 0.98 & 0.77 \\
\midrule
Model & Div & Ops & Cplex & Tot \\
\midrule
Codellama-7b  & 0.43 & 0.14 & 0.19 & 0.36\\
Llemma-7b & 0.81 & 0.21 & 0.21 & 0.51\\
Mistral-7b & 0.80 & 0.21 & 0.24 & 0.53 \\
Mixtral-8x7b-Instruct & 0.94 & 0.39 & 0.29 & 0.62\\
\bottomrule
\end{tabular}
\end{sc}
\end{footnotesize}
\caption{Accuracy for different LLMs evaluated on the synthetic dataset from elementary to complex skills and the overall accuracy. We observe that Mixtral-8x7b-Instruct is the best-performing model and that the more complex math skills are, the lower the accuracy across all models.}
\label{table:synthetic_accuracy}
\end{table}

\begin{figure*}
    \centering
    \includegraphics[scale=0.3]{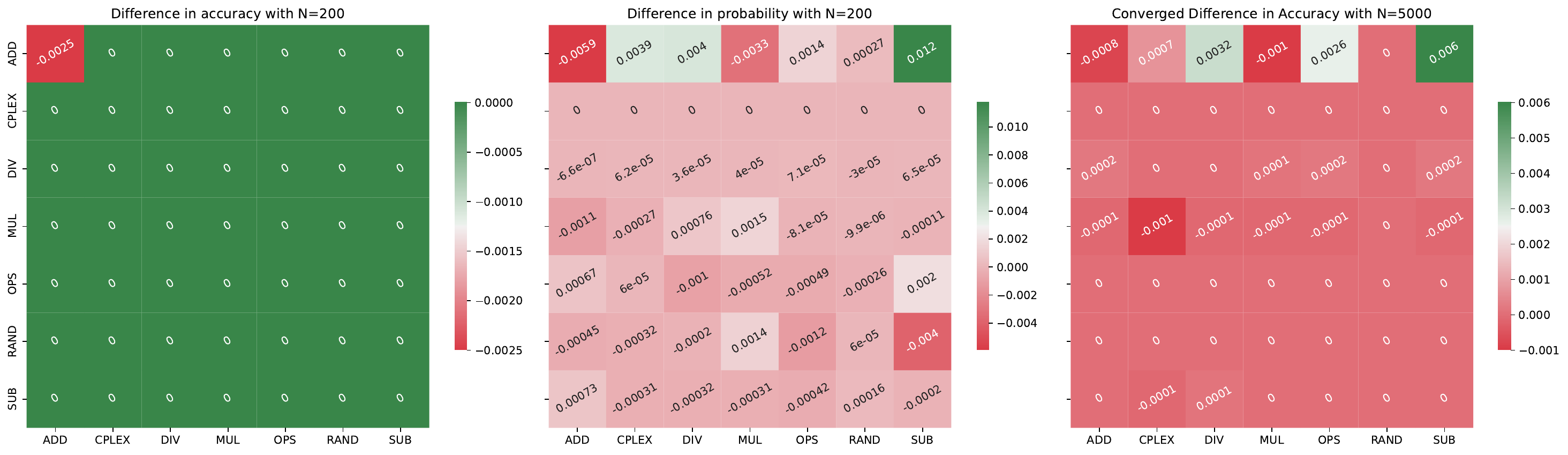}
    \caption{Illustration of sample efficiency of \textit{NTKEval} for Codellama-7b. The matrix shows either accuracy difference or difference in probability evaluated on row-specific skills between models instruction-tuned on column-specific skills and the base model. Left: Accuracy difference with $200$ generations per test question; Middle: Probability difference computed by \textit{NTKEval} with $200$ generations per test question; Right: Converged accuracy difference with $5000$ generations per test question. Green indicates positive and red indicates negative values. This shows that \textit{NTKEval} requires fewer generations compared to counting accuracy differences to capture changes in language model via instruction-tuning. \looseness=-1}
    \label{fig:sample_efficiency}
\end{figure*}

\begin{table}[t]
\centering
\begin{footnotesize}
\begin{sc}
\begin{tabular}{lccc}
\toprule
Model & GSM8K & KhanSkill & MATH \\
\midrule
Codellama-7b  & 0.10 & 0.15 & 0.06  \\
Llemma-7b & 0.39 & 0.31 & 0.15  \\
Mistral-7b & 0.38 & 0.28 & 0.11\\
\bottomrule
\end{tabular}
\end{sc}
\end{footnotesize}
\caption{Accuracy for different LLMs evaluated on KhanSkill, GSM8K and MATH. We observe KhanSkill has difficulty level easier than MATH and harder than GSM8K.}
\label{table:skill_accuracy}
\end{table}

\section{Experiments}
\label{sec:experiments}
\textbf{Model choices}
We evaluate our experiments on Code Llama 7b \cite{roziere2023code}, Llemma 7b \cite{azerbayev2023llemma} and either Mistral 7b \cite{jiang2023mistral} or Mixtral 8x7b Instruct \cite{jiang2024mixtral}. The choice of open-sourced models allows both inference and/or instruction-tuning done on a single GPU. We include a suite of LLMs tailored for code, mathematics, and SOTA general-purpose chat model in order to test specialized models' domain understanding. 

\textbf{Dataset Evaluation} Table \ref{table:synthetic_accuracy} records the accuracy for different LLMs evaluated on synthetic dataset from elementary to complex skills. We observe that the more complex skills (e.g. OPS and CPLEX) required to solve the problem the lower the accuracy across all models. Table \ref{table:skill_accuracy} reports the accuracy for different LLMs evaluated on KhanSkill dataset. We compare against standard benchmarks GSM8K \cite{cobbe2021gsm8k}, MATH \cite{hendrycksmath2021} to assess its difficulty level. We observe KhanSkill is easier than MATH and harder than GSM8K. In particular, Llemma-7b outperforms Mistral-7b on all three datasets in Table \ref{table:skill_accuracy}. \looseness=-1

\begin{figure*}
    \centering
    \includegraphics[scale=0.3]{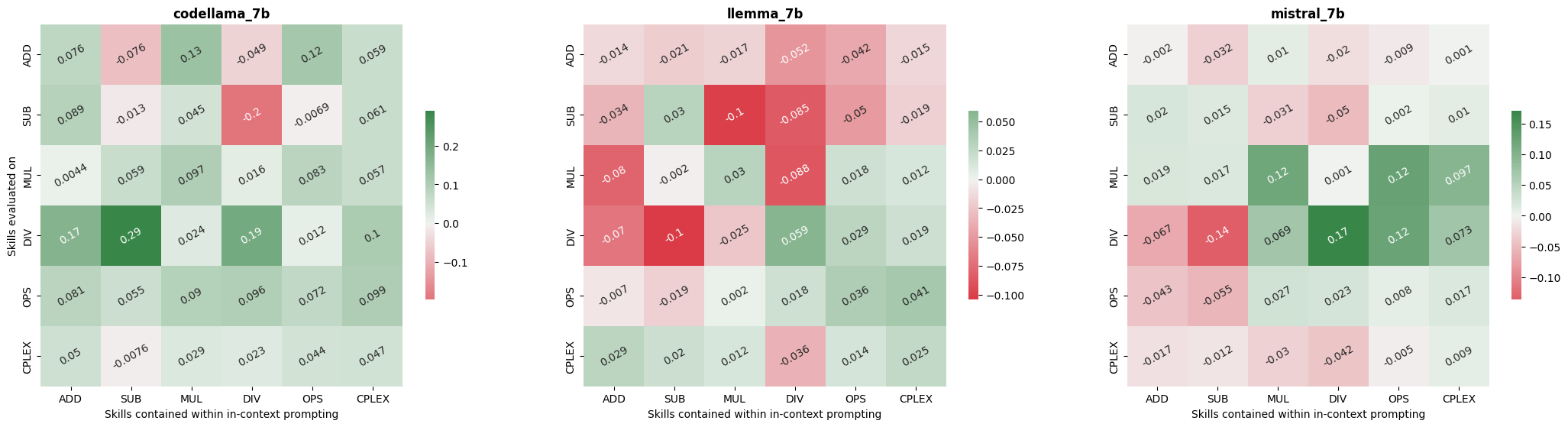}
    \caption{Accuracy difference between targeted skill prompting and standard prompting when in-context examples grouped by column-specified skills and evaluated on row-specified skills with base model as Codellama-7b (Left), Llemma-7b (Middle) and Mistral-7b (Right).  We observe Llemma-7b (LLM tailored for Mathematics) displays the most clear positive diagonal line, suggesting it is good at differentiating the targeted math skill
from the other relevant but misleading skills in ICL. Here green indicates positive values and red indicates negative values. }
    \label{fig:ICL_acc_matrix_synthetic2}
\end{figure*}

\subsection{\textit{NTKEval} Sample Efficiency}
\label{sec:NTKEval_sample_efficiency}
\textit{NTKEval} measures the difference in probability of generating correct solutions between a model trained on a skill-focused dataset and the base model. With sufficient generations, one expects the standard metric in reporting differences in accuracy between instruction-tuned and base models should converge to the difference in probability for generating correct solutions. Formally, for each QA pair $(x, y)$:
\begin{align}
    \text{NTKEval:} \quad &p_1(y|x) - p_0(y|x) \\
    \text{Accuracy:} \quad & \sum_{c_i \sim p_1(\cdot|x)} \mathbbm{1}_{y_i = y} - \sum_{c_i \sim p_0(\cdot|x)} \mathbbm{1}_{y_i = y}, 
\end{align}
where $c_i$ are $i$-th completion given the question $x$ sampled from fine-tuned model $p_1$ or base model $p_0$ and $y_i$ are solution extracted from the completion. 

As convergence is indifferent to the number of data points, we take the first $14$ data points in synthetic dataset, containing $2$ complete rounds of different skills. For each model, we observe converged accuracy differences with $5000$ generations per question at temperature $0.1$. We compute both \textit{NTKEval} and accuracy difference on the same dataset with $200$ generations at temperature $0.1$. Figure \ref{fig:sample_efficiency} shows (left) accuracy difference with $200$ generations (middle) \textit{NTKEval} with $200$ generations, and (right) converged accuracy difference with $5000$ generations. We observe with $200$ generations, the standard metric in counting accuracy does not capture changes in LM's probability distribution, i.e., it does not match the converged accuracy difference. The proposed method \textit{NTKEval} with only $200$ generations can reflect similarly in proportion to the converged accuracy difference. This suggests \textit{NTKEval} is a sample-efficient method compared to the standard metric in counting accuracy differences that allows one to understand how small changes in the language model's parameter space reflect in evaluation. Appendix \ref{appendix:sample_efficiency} details the experimental setup and includes figures that show such sample-efficiency feature holds true for all models considered. 

\textbf{Experiment setup}  

\textit{In-context learning}: standard prompting --- we consider 8-shot in-context examples, where examples are selected randomly from the training data; skill prompting --- we group 8-shot in-context examples that belong to the same category.

\textit{Instruction tuning}: base model --- we instruction-tuned LLM on a dataset that outputs random answers to reduce variability in probability changes unique to each LLM; skill-tuned model --- we instruction-tuned LLM on a dataset where questions belong to the same category.

Note, in both cases, the category can either be a math skill or a type of question format. Experiments compare changes in accuracy between skill prompting and standard prompting. Similarly, experiments compare changes in probability differences measured via \textit{NTKEval} between the skill-tuned model and base model. 
 \looseness=-1

\begin{figure*}
    \centering
    \includegraphics[scale=0.45]{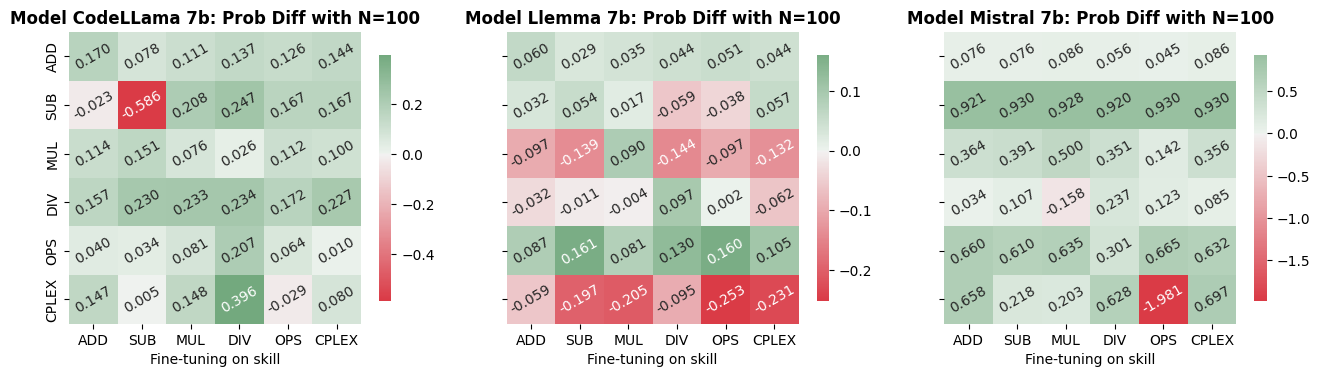}
    \caption{NTK matrix records changes in probability between the model instruction-tuned on column-specified skill and the base model evaluated on row-specified skill, where the base models are CodeLlama 7b (Left), Llemma-7b (Middle) and Mistral 7b (Right). Green indicates positive and red indicates negative values. We observe that instruction-tuning on column-specified skill datasets displays a positive diagonal line for the majority if not all skills, confirming that training on a targeted skill improves using the skill at test time. }
    \label{fig:NTK_kernel_temp01_ngen_100}
\end{figure*}

\subsection{Do LLMs answer math problems based on deep or surface-level
structures?}
\label{sec:deep_shallow}
We evaluate LLMs' ability to learn beyond pattern matching by asking: do LLMs outperform when seeing examples that contain core math skills versus those that contain superficial consistent question formats? \looseness=-1

We hypothesize that if LLMs predict based on pattern matching, then changes in performance would be negligible when prompting / instruction-tuning based on deep structures versus that based on surface structures. 

\textit{In-context learning} We select in-context examples that have the same category as the test question. For instance, we select in-context examples that test addition for addition problems in the test set; we select examples that have style `symbolic' for symbolic problems in the test set. The categories we considered are four elementary math skills and four presentation formats. 

\textit{Instruction-tuning} Similar to in-context learning, for each deep and surface structure, the training and test datasets share the same structure, where we train and evaluate the model on its corresponding test dataset. 

\begin{table}[t]
\begin{center}
\begin{footnotesize}
\begin{sc}
\begin{tabular}{lcc}
\toprule
 Model & Deep  & Surface \\
\midrule
Codellama-7b  &  0.008 & $-$0.071 \\
Llemma-7b &  0.022 & $-$0.026 \\
Mistral-7b &  0.024 & 0.037 \\

\bottomrule
\end{tabular}
\end{sc}
\end{footnotesize}
\end{center}
\vskip -0.15in
\caption{Mean accuracy differences compared to the standard prompting when in-context examples share the same structure as the test question. The results are averaged over different types of deep and surface structures. We observe that Code LLama 7b and Llemma-7b (LLM for code and math) show a strong relative improvement when prompted with deep math structures, indicating a level of domain understanding.}
\label{table:ICL_deep_shallow}
\end{table}

Table \ref{table:ICL_deep_shallow} reports accuracy differences between prompting based on structure and standard prompting evaluated on test problems that belong to the same structure. The results are averaged across four deep structures and four surface structures, respectively. We observe LLMs tailored for code and mathematics both show a strong improvement in accuracy when seeing examples related to deep structures versus merely seeing examples related to surface structures. This suggests those LLMs output answers not merely based on pattern matching in question formats but also develops a deeper understanding of mathematical skill to improve performance based on seeing more relevant in-context examples. We also observe that Mistral 7b is the best-performing model as irrespective of deep or surface prompting, the model shows the biggest improvement compared to others. However, Mistral 7b shows a decrease in performance when seeing examples related to deep structures compared to surface structures. This suggests Mistral 7b though better in performance when seeing relevant problems does not show an understanding of domain-specific structures. As expected, LLM tailored for mathematics (Llemma-7b) exhibits the best understanding (i.e., the largest relative improvement) with deep mathematical structures.  \looseness=-1

\begin{table}[t]
\begin{center}
\begin{footnotesize}
\begin{sc}
\begin{tabular}{lcc}
\toprule
 Model & Deep  & Surface \\
\midrule
Codellama-7b  &  0.25 & 0.21 \\
Llemma-7b &  0.29 & 0.33 \\
Mistral-7b &  -0.75 & -0.12 \\

\bottomrule
\end{tabular}
\end{sc}
\end{footnotesize}
\end{center}
\vskip -0.15in
\caption{Mean probability differences compared to the base model when training dataset shares the same structure as the test dataset. The results are averaged over different types of deep and surface structures. We observe there is a decline if not negligible relative improvement for LLMs training on data share deep structures, indicating a lack of domain understanding. \looseness=-1 
}
\label{table:FT_deep_shallow}
\end{table}

Table \ref{table:FT_deep_shallow} reports the average probability difference between model instruction tuned on one structure and the base model evaluated on test problems that belong to the same structure. The results are averaged over four deep structures and four surface structures, respectively. We observe that Llemma-7b and Mistral-7b demonstrate a relative decrease in probability improvement when training on deep structures compared to that of surface structures, and Codellama-7b also exhibits a negligible improvement in scale. This suggests instruction tuning on datasets containing singleton skill does not help language models to recognize deep mathematical structures. Appendix \ref{appendix:deep_shallow} records detailed experimental setup and structure-stratified results.  \looseness=-1

\subsection{Can LLM understand different math skills?}
\label{sec:elementary_to_complex}
We assess LLMs' learning-to-learn ability when seeing relevant examples grouped by skills affecting utilizing the target skill and different skills in test time. 

\textit{Setup} For each elementary to complex skill (Figure \ref{fig:sample_synthetic_questions} Left), we prompt LLMs with 8 in-context examples grouped by skill or instruction-tune the LLM with skill-specified data and evaluate the model on the overall test dataset stratified by skills. 

Figure \ref{fig:ICL_acc_matrix_synthetic2} shows ICL matrix that records changes in accuracy compared to standard prompting when in-context examples are grouped as column-specified skills and evaluated on row skills.  We observe all models display a positive diagonal line for the majority of skills. This suggests giving relevant examples during inference time, LLMs in general are fast in learning, with performance improvement on the corresponding examples during test time. As expected for LLM tailored for mathematics, Llemma-7b exhibits the most clear positive diagonal line, suggesting the clear differentiation from the targeted mathematical skill with the other relevant but misleading skills. \looseness=-1

Figure \ref{fig:NTK_kernel_temp01_ngen_100} shows NTK matrix that records changes in probability of generating correct solutions compared to base model when models are instruction-tuned on column-specified skill dataset and evaluated on questions with row-specified skills. All models display a positive diagonal line for the majority, if not all skills. This suggests language model via tuning is also fast to adapt to using that skill during test time. 

\subsubsection{Can LLM distinguish targeted skill from off-diagonal skills?}
To distinguish whether the ability for fast adaptation (i.e., improvement) observed in Figure \ref{fig:ICL_acc_matrix_synthetic2} and \ref{fig:NTK_kernel_temp01_ngen_100} is not due to shared factors (e.g., formatting), we compare the scale of improvement when prompting / instruction-tuning between targeted and off-diagonal skills. Targeted skill refers to the skill that is the same as the test question. Off-diagonal skills refer to skills related to but distinct from the test questions. For example, if the target question requires skill `addition', then off-diagonal skills include but are not limited to `subtraction'.

\begin{figure}[h]
    \centering
    \begin{subfigure}{}
    \includegraphics[width=\columnwidth]{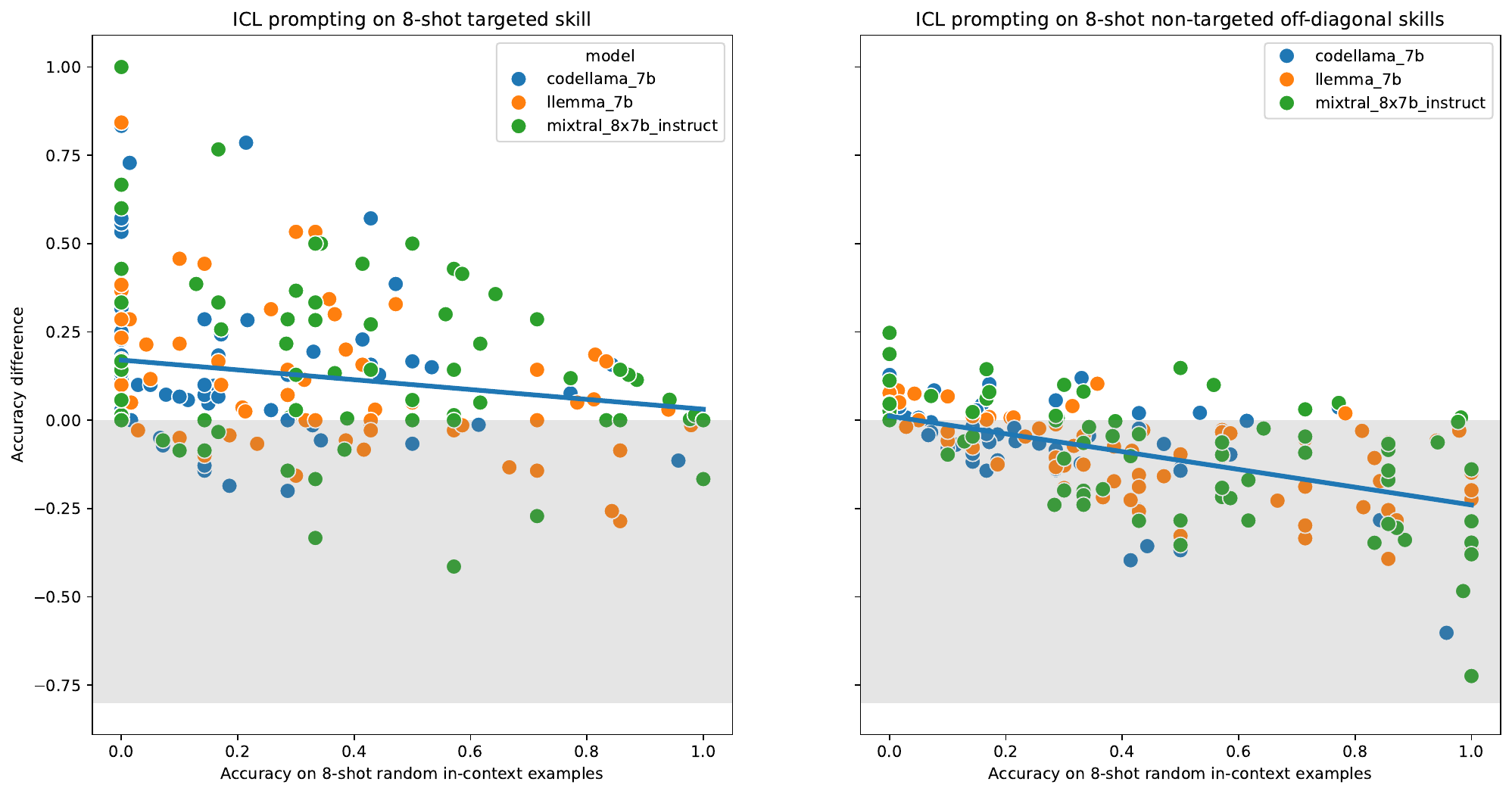}
    \end{subfigure}
    \begin{subfigure}{}
    \includegraphics[width=\columnwidth]{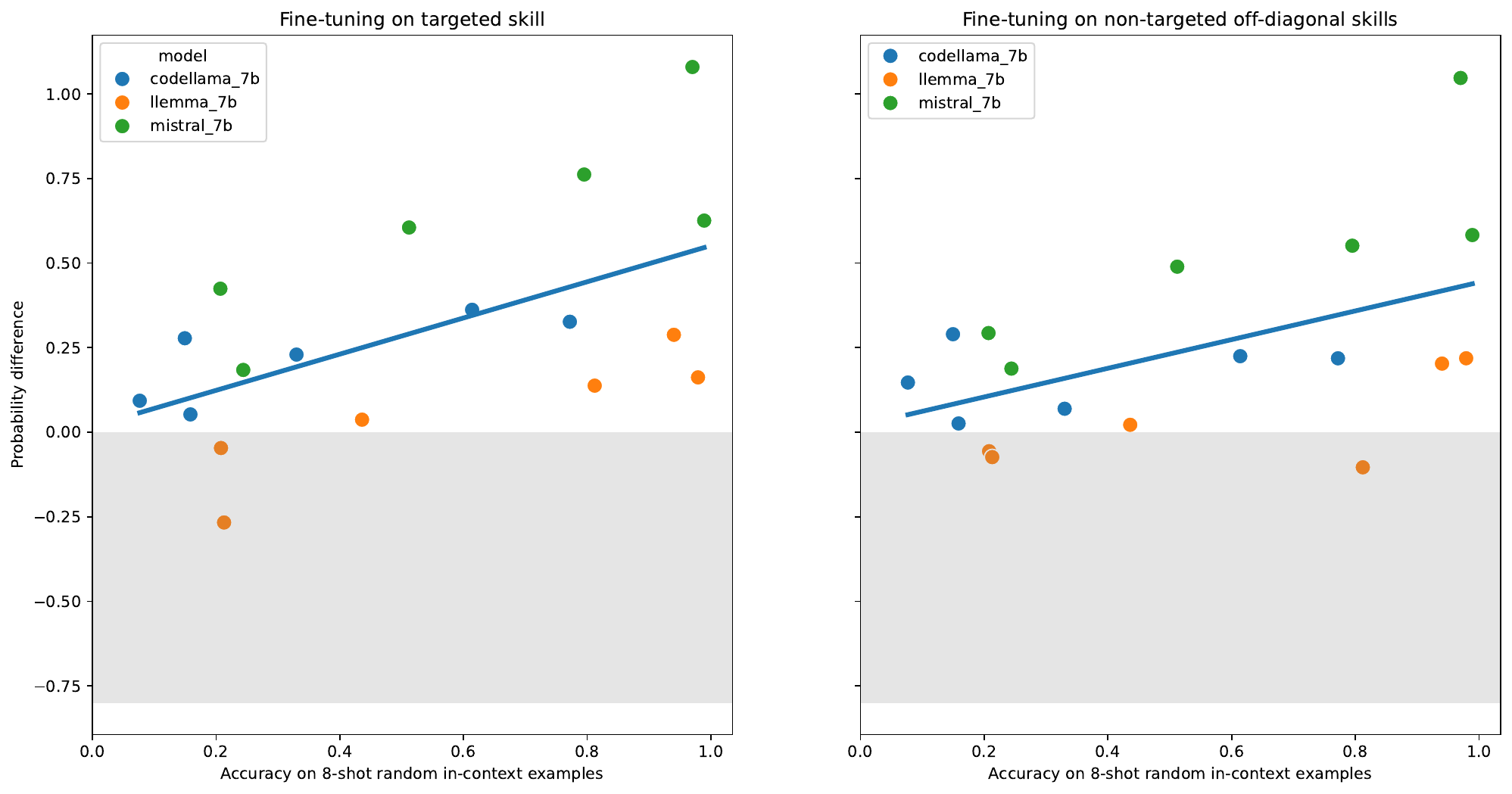}
    \end{subfigure}
    \vspace{-0.3cm}
    \caption{Top: ICL accuracy difference on targeted (top left) and off-diagonal (top right) skill prompting compared to standard prompting; Bottom: Average difference in probability of generating correct solutions when instruction-tuning on targeted (left) and off-diagonal (right) skills compared to the base model. The x-axis displays the difficulty level of individual skills (measured by accuracy under 8-shot random in-context examples). In-context learning is able to differentiate the targeted skill from off-diagonal skills through the clear relative accuracy improvement, whereas instruction-tuning shows similar performance improvement irrespective of training skills. \looseness=-1}
    \label{fig:targetd_vs_off_diagonals}
\end{figure}

Figure \ref{fig:targetd_vs_off_diagonals} (Top) shows average changes in accuracy when performing targeted skill prompting (top left) and off-diagonal skill prompting (top right) compared to the standard prompting on both synthetic and KhanSkill dataset. Figure \ref{fig:targetd_vs_off_diagonals} (Bottom) shows average changes in the probability of generating correct solutions when instruction-tuning on targeted skill (bottom left) and off-diagonal skills (bottom right) compared to the base model on a synthetic dataset. The x-axis shows skill difficulty levels measured by accuracy under 8-shot random in-context examples for each corresponding model. \looseness=-1
 
\textbf{Observations} Figure \ref{fig:targetd_vs_off_diagonals} (Top) shows clear advantages in relative accuracy improvement, whereas (Bottom) shows similar performance changes between seeing targeted skill (Left) versus off-diagonal skills (Right). This suggests the qualitative improvements observed in in-context learning (Figure \ref{fig:ICL_acc_matrix_synthetic2}) are not due to irrelevant factors but stem from the differentiation between different mathematical skills. 
By contrast, instruction-tuning does not demonstrate relative advantages when trained on the targeted skill versus off-diagonal skills. This suggests the qualitative performance improvement (Figure \ref{fig:NTK_kernel_temp01_ngen_100}) is driven not by differentiation on relevant skills but rather by common features shared across all tested skills (e.g., presentation format).  \looseness=-1 

\section{Conclusion}
We proposed a method \textit{NTKEval} to extend computations of neural tangent kernel beyond regression tasks to language model's outputs as free-form generations (Section \ref{sec:NTKEval}). Figure \ref{fig:sample_efficiency} demonstrates the sample-efficiency of \textit{NTKEval} across all models. We next utilize \textit{NTKEval} and investigate the capability of LLMs to answer math problems beyond pattern matching through query after in-context learning and instruction-tuning. Experiment results show in-context learning differentiates deep structures from surface structures (Table \ref{table:ICL_deep_shallow}), by contrast instruction-tuning does not (Table \ref{table:FT_deep_shallow}). Both ICL and IT are capable of learning to learn (Fig. \ref{fig:ICL_acc_matrix_synthetic2} and \ref{fig:NTK_kernel_temp01_ngen_100}), but in-context learning identifies targeted math skills from the others (Figure \ref{fig:targetd_vs_off_diagonals} Top), whereas instruction-tuning does not (Figure \ref{fig:targetd_vs_off_diagonals} Bottom). Overall, we find that ICL exhibits domain understanding, whereas certain instruction-tuning leads to similar performance change irrespective of training on different data. \looseness=-1

We only considered QA data rather than open-ended texts. We investigated datasets grouped by one skill rather than a mixture of diverse data. Through investigating whether and what method elicits LLM's domain understanding, we hope to help design better and more transparent scientific assistants.
 
\section*{Broader Impact}
This paper presents work whose goal is to advance the field of Machine Learning. There are many potential societal consequences of our work, none which we feel must be specifically highlighted here.
\bibliography{main}
\bibliographystyle{icml2023}

\newpage
\appendix
\onecolumn
\section{Dataset}
\label{sec:appendix_dataset}
KhanSkill dataset consists of questions generated from \url{https://github.com/Khan/khan-exercises}. The questions consist of $93$ skills with $20$ questions per skill. Figure \ref{fig:khanskill_skills_sample_questions} includes a list of sample skill names and a sample data point in the dataset.  Khan Exercises is licensed under Copyright (c) 2015 Khan Academy, with the exercise framework is MIT licensed and the exercises are under a Creative Commons by-nc-sa license. Note we later find that AMPS \cite{hendrycks2020measuring} contains questions also generated under the same github, but due to different checkpoint generations, KhanSkill and AMPS contain both overlapping and distinct skills and questions. \looseness =-1

\begin{figure}[h]
\begin{minipage}{1\textwidth}
\begin{tcolorbox}[colback=gray!5!white,colframe=gray!75!black,title=Example skills involved in KhanSkill Dataset, raster equal height=rows]
divisibility, dividing\_fractions\_0.5,
converting\_repeating\_decimals\_to\_fractions\_2,  \\
dividing\_fractions\_alternative, units, 
exponents\_2\_alternative, chain\_rule\_1 \\
adding\_and\_subtracting\_fractions, markup\_and\_commission\_word\_problems \\
product\_rule, quotient\_rule, expressing\_ratios\_as\_fractions, \\properties\_of\_numbers\_2, 
multiplying\_fractions\_by\_integers, and more
\end{tcolorbox}
\end{minipage}
\vspace{0.3cm}
\begin{minipage}{1\textwidth}
\begin{tcolorbox}[colback=gray!5!white,colframe=gray!75!black,title=Sample Question in KhanSkill Dataset, raster equal height=rows]
Skill: dividing\_fractions\_0.5 \\ 
Question: \\
Reduce to lowest terms: $\displaystyle \frac{8}{5} \div \frac{1}{3} = {?}$ \\
Answer: \\
Turn into a multiplication problem: $\displaystyle {?} = \frac{8}{5} \times \frac{3}{1}$. Combine into one fraction: 
$\displaystyle {?} = \frac{8 \times 3}{5 \times 1}$. Simplify: 
$\displaystyle {?} = \dfrac{24}{5}$ \\
\# $4.8$
\end{tcolorbox}
\end{minipage}
\caption{Sample skills (top) and sample question (bottom) included in KhanSkill Dataset.}
\label{fig:khanskill_skills_sample_questions}
\end{figure}

\section{Sample Efficiency}
\label{appendix:sample_efficiency}
\begin{figure}
\centering
\begin{subfigure}{}
       \includegraphics[width=.8\linewidth]{figures/codellama_7b_convergence.pdf}
   \caption{Illustration on sample efficiency of \textit{NTKEval} for Codellama-7b. Matrix shows either accuracy difference or difference in probability of generating correct solutions evaluated on row-specific skills between models instruction-tuned on column-specific skills and the base model. Left: Accuracy difference with $200$ generations per test question; Middle: Probability difference computed by \textit{NTKEval} with $200$ generations per test question; Right: Converged accuracy difference with $5000$ generations per test question. Green indicates positive values and red indicates negative values.}
\label{fig:codellama_7b_convergence}
\end{subfigure}
\vspace{.3cm}
\begin{subfigure}{}
    \includegraphics[width=.8\linewidth]{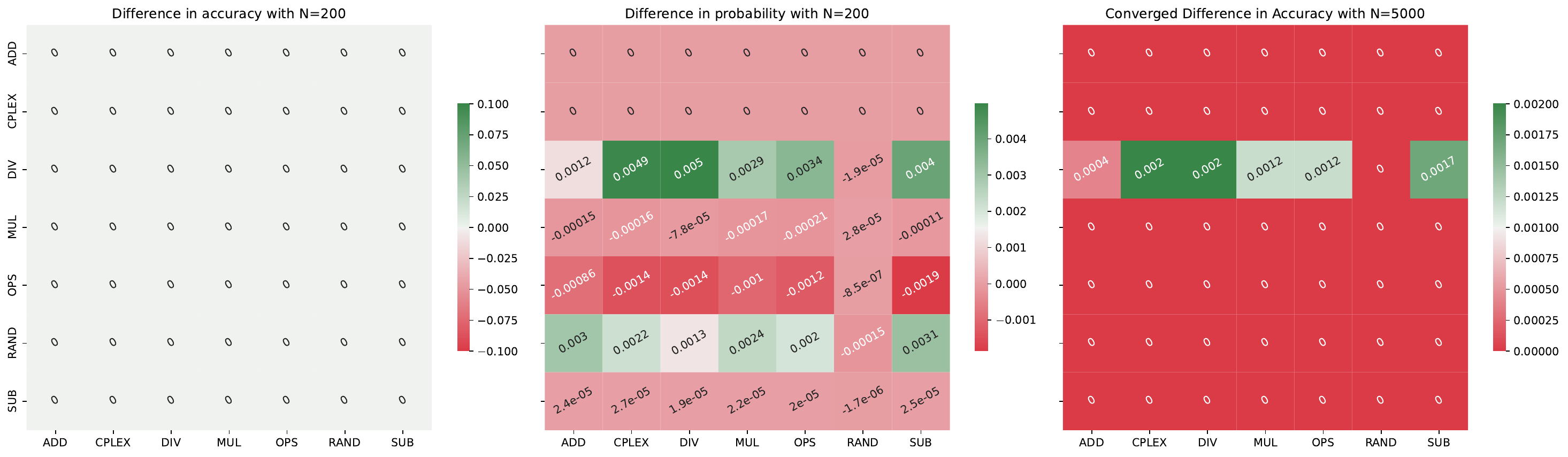}
\caption{Illustration on sample efficiency of \textit{NTKEval} for Llemma-7b. Matrix shows either accuracy difference or difference in probability of generating correct solutions evaluated on row-specific skills between models instruction-tuned on column-specific skills and the base model. Left: Accuracy difference with $200$ generations per test question; Middle: Probability difference computed by \textit{NTKEval} with $200$ generations per test question; Right: Converged accuracy difference with $5000$ generations per test question. Green indicates positive values and red indicates negative values.}
\label{fig:llemma_7b_convergence}
\end{subfigure}
\vspace{.3cm}
\begin{subfigure}{}
\includegraphics[width=.8\linewidth]{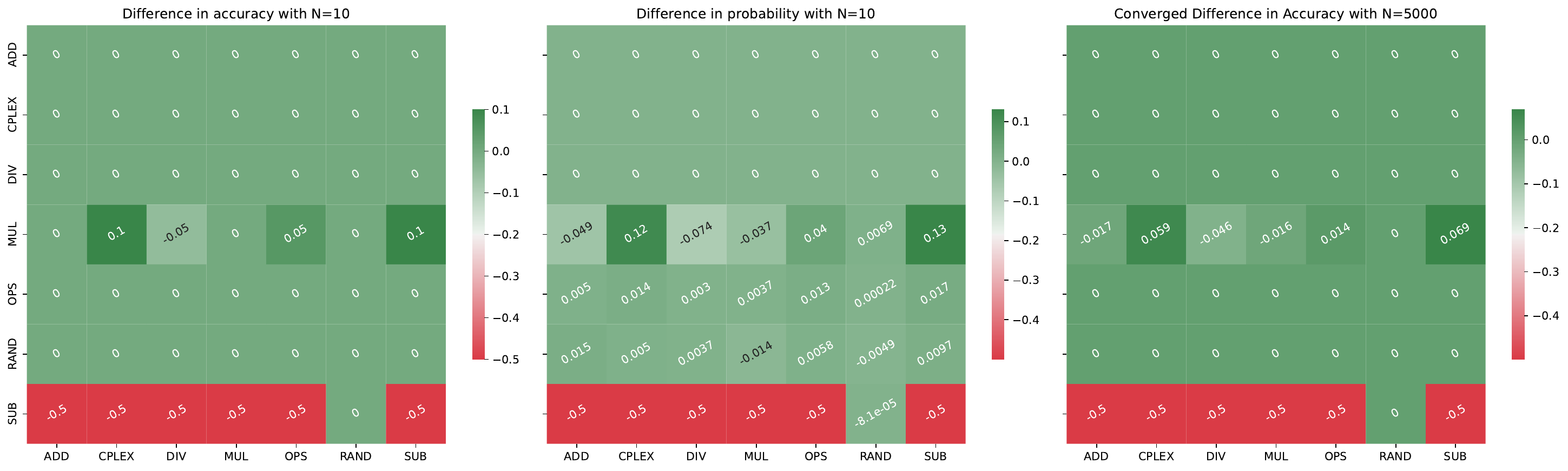}
\caption{Illustration on sample efficiency of \textit{NTKEval} for Mistral-7b. Matrix shows either accuracy difference or difference in probability of generating correct solutions evaluated on row-specific skills between models instruction-tuned on column-specific skills and the base model. Left: Accuracy difference with $10$ generations per test question; Middle: Probability difference computed by \textit{NTKEval} with $10$ generations per test question; Right: Converged accuracy difference with $5000$ generations per test question. Green indicates positive values and red indicates negative values.}
\label{fig:mistral_7b_convergence} 
\end{subfigure}
\label{fig:convergence}
\end{figure}

Experiment setup: The training dataset consists of $2000$ questions that share the same skill as column-specified. We perform stochastic gradient descent on one GPU with batch size $8$, learning rate \num{2e-5} and warm up ratio $0.03$ with cosine learning rate scheduler for one epoch. We compute differences in accuracy and probability with respect to the model instruction-tuned on skill-focused dataset and the model instruction-tuned on random dataset (i.e., a dataset that inherits the same question format but outputs random integers from $1$ to $1000$). We evaluate the models on the overall test dataset which consists $100$ questions per skill and visualize the results with skill stratification. Figure \ref{fig:codellama_7b_convergence} - \ref{fig:mistral_7b_convergence} demonstrate the sample efficiency advantage brought by \textit{NTKEval} holds true across three open-models the paper tested on.

\section{Can LLM recognize deep vs. surface structures?}
\label{appendix:deep_shallow}
\begin{figure}
    \centering
\begin{subfigure}{}
   \includegraphics[width=\linewidth]{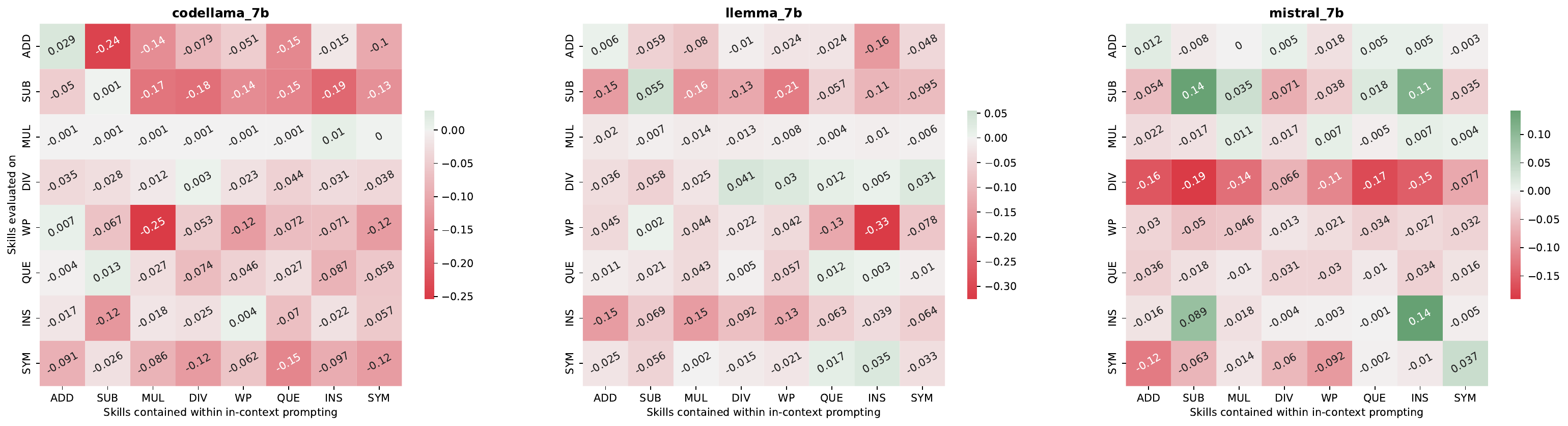}
   \caption{Accuracy difference between targeted skill prompting and standard prompting when in-context examples grouped by column-specified skills and evaluated on row-specified skills with the base model as Codellama-7b (Left), Llemma-7b (Middle) and Mistral-7b (Right).  Green indicates positive values and red indicates negative values.}
   \label{fig:ICL_acc_matrix_synthetic3}
\end{subfigure}
\begin{subfigure}{}
   \includegraphics[width=\linewidth]{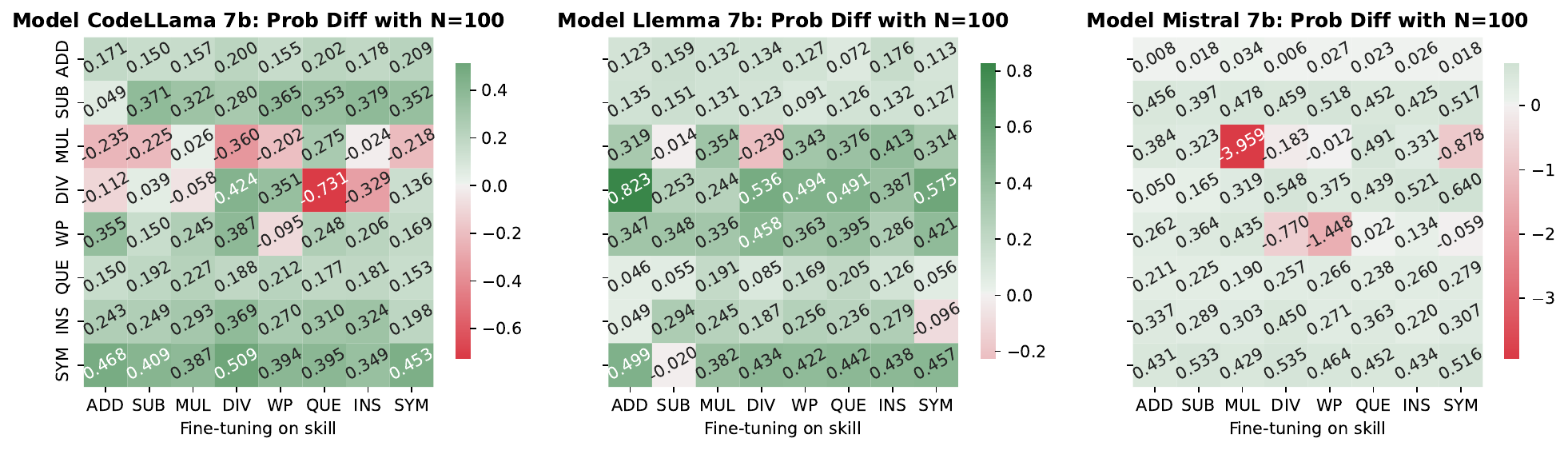}
   \caption{NTK matrix records changes in probability between the model instruction-tuned on column-specified skill and the base model evaluated on row-specified skill, where the base models are CodeLlama 7b (Left), Llemma-7b (Middle) and Mistral 7b (Right). Green indicates positive and red indicates negative values.}
   \label{fig:NTK_matrix_synthetic3} 
\end{subfigure}
    \label{fig:appendix_deep_shallow}
\end{figure}

In-context learning: all generations are sampled from temperature $0.1$ with $10$ generations per input. Instruction tuning: We train via stochastic gradient descent with learning rate \num{2e-3} for one epoch at batch size $8$ with $0.03$ warm-up ratio and cosine learning rate scheduler. Figure \ref{fig:ICL_acc_matrix_synthetic3} shows the accuracy matrix via ICL across four deep structures and four presentation formats. Figure \ref{fig:NTK_matrix_synthetic3} shows the NTK matrix measured via changes in probability after training across four deep structures and presentation formats and evaluated on a skill-stratified test set. 

\section{Can LLMs understand different math skills}
Figure \ref{fig:icl_template} shows the template of in-context learning instruction. We trained on $2000$ questions that share the same skill with SGD at batch-size $8$, learning rate \num{2e-3} and warm-up ratio $0.03$ with cosine learning rate scheduler. Predictions are sampled at temperature $1$ with $100$ generations per question for evaluation on instruction-tuned models.  \looseness=-1

\begin{figure}
\begin{minipage}{\textwidth}
\begin{tcolorbox}[colback=gray!5!white,colframe=gray!75!black,title=In-context learning template]
Below is an instruction that describes a task. Write a response that appropriately completes the request. \\
Instruction: What is 712 - 486 ? \\
Response: 226 \\ \\
Instruction: What is 434 - 363 ? \\ 
Response: 71 \\ \\
\ldots and more (8 examples in total)\\

Instruction: What is 10*84-25-(0 / (91 + 47))? \\
Response:
\end{tcolorbox}
\caption{In-context learning template when showing 8 in-context examples using the skill `subtraction' and evaluating on test question requires skill `complex'.}
\label{fig:icl_template}

\end{minipage}
\end{figure}

\end{document}